\documentclass[10pt,twocolumn,letterpaper]{article}

\usepackage{iccv}
\usepackage{times}
\usepackage{epsfig}
\usepackage{graphicx}
\usepackage{amsmath}
\usepackage{amssymb}

\usepackage{bm}
\usepackage{algorithm}
\usepackage{algorithmic}
\usepackage{multirow}
\usepackage{bbding}

\usepackage[accsupp]{axessibility}  

\usepackage[pagebackref=true,breaklinks=true,letterpaper=true,colorlinks,bookmarks=false]{hyperref}

\iccvfinalcopy 

\def\iccvPaperID{6199} 

\begin{document}

\title{SUNet: Symmetric Undistortion Network for Rolling Shutter Correction}

\author{Bin Fan ~~~~~~~~~~~~~ Yuchao Dai\thanks{Corresponding author} ~~~~~~~~~~~~~ Mingyi He\\
	School of Electronics and Information, Northwestern Polytechnical University, Xi'an, China\\
	{\tt\small binfan@mail.nwpu.edu.cn, daiyuchao@nwpu.edu.cn, myhe@nwpu.edu.cn}
}

\maketitle

\begin{abstract}
   The vast majority of modern consumer-grade cameras employ a rolling shutter mechanism, leading to image distortions if the camera moves during image acquisition. In this paper, we present a novel deep network to solve the generic rolling shutter correction problem with two consecutive frames. Our pipeline is symmetrically designed to predict the global shutter image corresponding to the intermediate time of these two frames, which is difficult for existing methods because it corresponds to a camera pose that differs most from the two frames. First, two time-symmetric dense undistortion flows are estimated by using well-established principles: pyramidal construction, warping, and cost volume processing. Then, both rolling shutter images are warped into a common global shutter one in the feature space, respectively. Finally, a symmetric consistency constraint is constructed in the image decoder to effectively aggregate the contextual cues of two rolling shutter images, thereby recovering the high-quality global shutter image. Extensive experiments with both synthetic and real data from public benchmarks demonstrate the superiority of our proposed approach over the state-of-the-art methods.
\end{abstract}

\vspace{-3.3mm}
\section{Introduction} \label{sec:Introduction}
\begin{figure}[!t]
	\centering
	\includegraphics[width=0.49\textwidth]{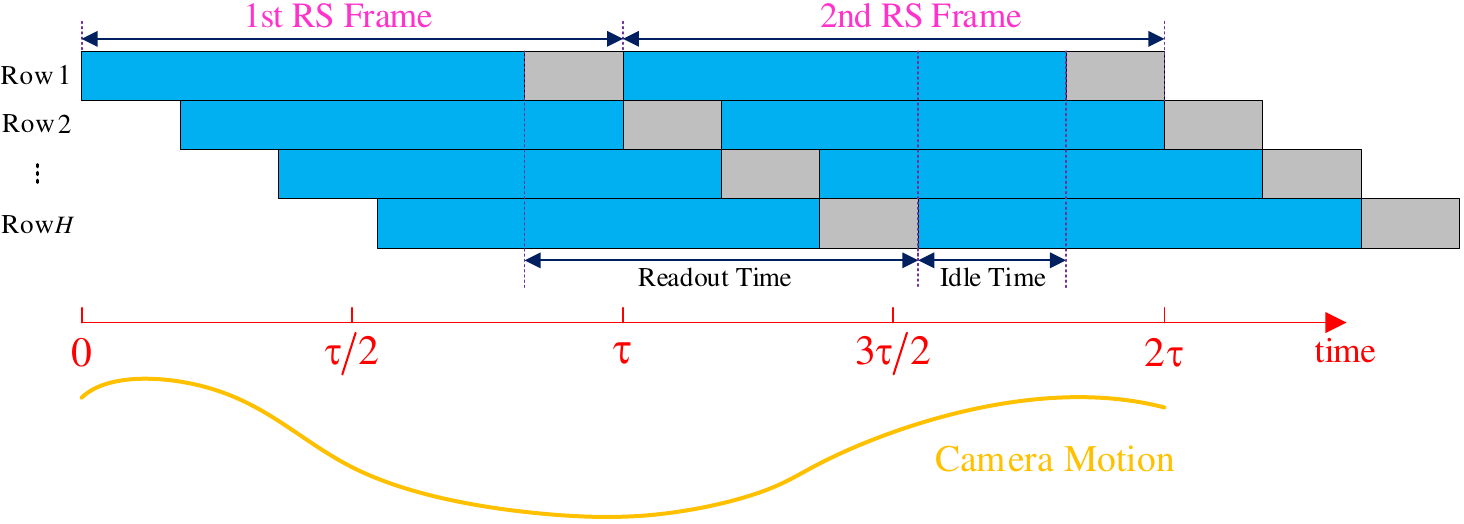}
	\caption{Illustration of the exposure mechanism of the RS camera over two consecutive frames. Assume the sensor exposure is instantaneous and each frame time is $\tau$. The sensor is exposed and readout row by row, resulting in the scanline-varying camera poses. Our approach aims to restore the time-centered GS image that corresponds to the intermediate time $\tau$ of these two frames.\label{fig:RS_mechanism}}
\end{figure}

Popular low-budget commercial cameras are generally built upon CMOS sensors due to their low cost and simplicity in design.
Most CMOS cameras employ a rolling shutter (RS) mechanism.
Different from the global shutter (GS) camera that exposes all pixels at the same time, the RS camera is exposed in a row-wise manner from top to bottom. Thus, the images and videos captured by a moving RS camera will have the RS effect (\emph{e.g.}, stretch, wobble).
We provide an illustration in Fig.~\ref{fig:RS_mechanism} while assuming the camera has a frame time of $\tau$. 
The high dynamic sampling characteristics of the RS mechanism are regarded as an advantage rather than a disadvantage \cite{ait2006simultaneous}, but simply ignoring the RS effects in 3D geometric vision applications may lead to erroneous, undesirable and distorted results \cite{dai2016rolling,lao2020rolling,albl2015R6P,fan2021rsdpsnet}.
To mitigate or eliminate the RS effect which is increasingly becoming a nuisance in photography, a well-known RS correction problem has therefore attracted more and more attention \cite{zhuang2017rolling,zhuang2019learning,liu2020deep}.
This challenging problem aims at recovering the GS image corresponding to the camera pose at a specific exposure time (\emph{e.g.}, $0$ or $\tau$ in Fig.~\ref{fig:RS_mechanism}).

\begin{figure*}[!t]
	\centering
	\includegraphics[width=0.937\textwidth]{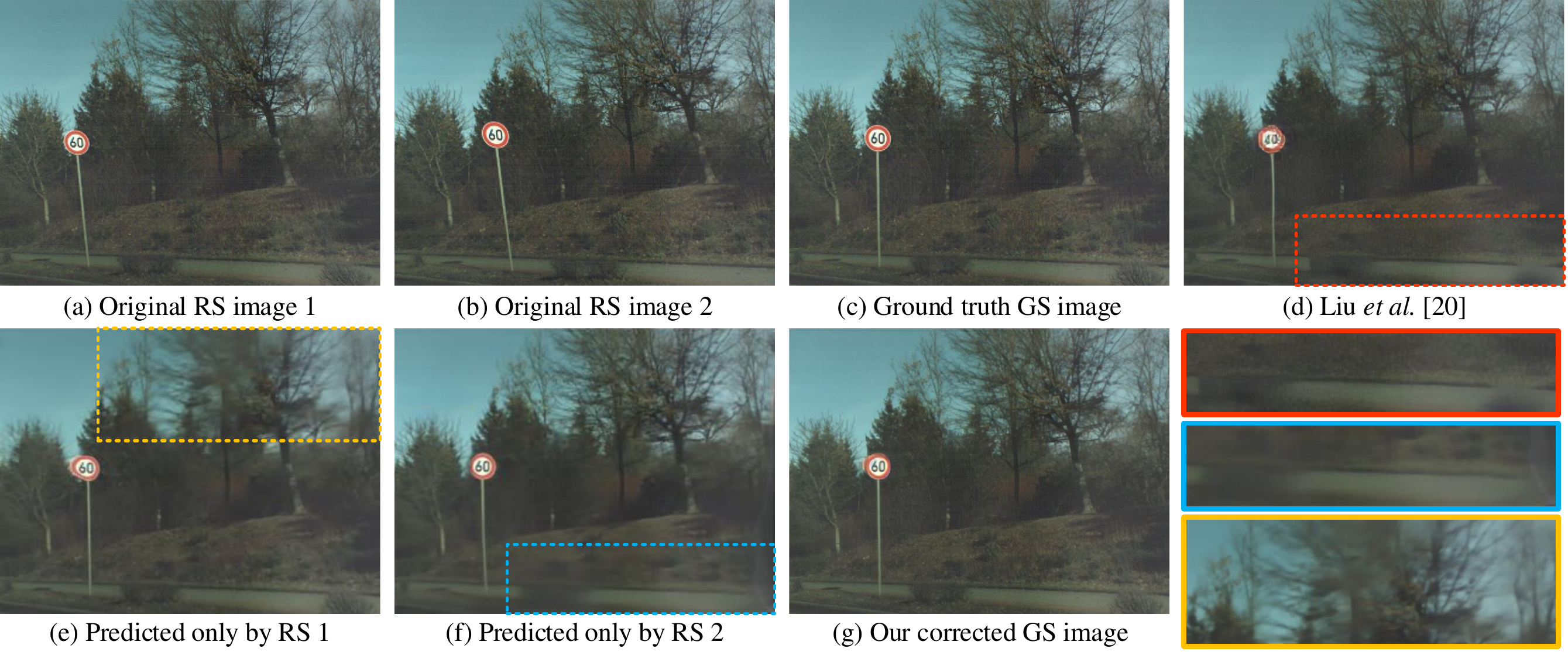}\vspace{-0.45mm}
	\caption{An example of RS correction from two consecutive frames. (a-b) Input two consecutive RS images; (c) Ground truth GS image corresponding to the middle time of two consecutive frames; 
		(d) GS image predicted by the state-of-the-art method \cite{liu2020deep}; (e-g) The forward, backward, and target GS images predicted from only the first RS image, only the second RS image, and the combined two RS images (our pipeline), respectively.
		The first and second RS images have different contributions to the upper and lower regions of the corresponding GS image which can be seen between yellow and blue boxes.
		Our proposed approach makes use of this property for RS correction. \label{fig:overview}}
	\vspace{-0.47cm}
\end{figure*}

Existing works on RS correction generally fall into one of two categories: single-frame-based or multi-frame-based.
It is an ill-posed problem in the case of single-frame-based RS correction, where additional prior assumptions need to be formulated explicitly \cite{rengarajan2016bows,lao2018robust,purkait2017rolling,Purkait2018Minimal} or implicitly
\cite{rengarajan2017unrolling,zhuang2019learning}.
Following the previous studies \cite{zhuang2017rolling,zhuang2020homography,liu2020deep,fan2021rssr} in reducing the ill-posedness, we also focus on the well-posed generic RS correction problem from two consecutive frames of a video. 
In particular, since RS cameras are usually time-synchronized with other sensors (\emph{e.g.}, GS camera, IMU, etc.) in hardware by referring to the first scanline time \cite{schubert2019rolling,hedborg2012rolling}, we deal with a corresponding challenge of correcting the RS images to the GS image corresponding to the exposure time of the first scanline of the second frame (\emph{i.e.}, the intermediate time $\tau$ of these two frames).
This is of both theoretical interest and great practical importance.

To this end, classical two-frame-based methods \cite{zhuang2017rolling,zhuang2020homography} heavily rely on specific RS motion models and require nontrivial iterative optimizations, which limits their use in time-constrained applications.
Although the state-of-the-art learning-based method \cite{liu2020deep} has achieved promising results in recovering the GS image corresponding to the middle time $\frac{3\tau}{2}$ of the second frame, it is prone to fail to predict a plausible GS image at time $\tau$, where many image texture details cannot be well restored (\cf Fig.~\ref{fig:overview}(d)).
Since the GS image at time $\frac{3\tau}{2}$ and the second RS image have close content, their asymmetric network only exploits the limited information of the second RS image, but ignores the contextual aggregation of both RS frames.

Recently, various image-to-image translation problems have been tackled with the deep learning pipelines (\emph{e.g.}, image correction \cite{liu2020deep}, optical flow estimation \cite{sun2018pwc}, depth estimation \cite{mayer2016large}, video deblurring \cite{su2017deep}, and image super-resolution \cite{lim2017enhanced}, etc.). 
However, it still poses a challenge to learn to produce a satisfactory GS image corresponding to the intermediate time $\tau$ of two consecutive frames.
Its complexities mainly lie in the following facts: 
\textbf{1}) Different from the local neighbors that can provide sufficient description in general image-to-image translation problem, a pixel of the target GS image may not be in the neighboring pixel of its corresponding RS images, depending on the type of motion, the 3D structure, and the scanline time;
\textbf{2}) We observe in Fig.~\ref{fig:overview}(e)\&(f) that the first and second RS images contribute greatly to the lower and upper parts of the corresponding time-centered GS image, respectively.
This is expected because they are closer in time and thus share more similar camera poses.
Consequently, the centered exposure time $\tau$ corresponds to the most unique camera pose that deviates greatly from both RS frames, such that only the cues from a single RS image may result in the lack of details of the corrected GS image. 
Therefore, motivated by these two insights, we propose a novel symmetric undistortion network architecture to rectify the geometric inaccuracies induced by the RS effect from two consecutive RS images.
The network can benefit from the overall exploitation and aggregation of contextual information, where the time-centered GS image can be reconstructed by using an adaptive fusion scheme, as illustrated in Fig.~\ref{fig:overview}(g).

Our network takes two consecutive RS images as input and predicts the corrected GS image corresponding to the intermediate time of these two frames. To essentially exploit and merge contextual information across both RS images, our pipeline is symmetrically constructed. 
It consists of two main processes: a PWC (pyramid, warping, and context-aware cost volume)-based undistortion flow estimator and a time-centered GS image decoder.
The success of the classic PWC-Net framework \cite{sun2018pwc} in aggregating multi-scale context information inspires the construction of our undistortion flow estimator, which is to estimate the pixel-wise undistortion flows of the first and second RS images.
Note that the \emph{context-aware cost volume} we construct can effectively promote contextual consistency at different scales.
Then, the pixel-wise undistortion flows are used to warp the learned image features to their corresponding GS counterparts.
Finally, we develop a time-centered GS image decoder to further align the contextual cues of their warped features and convert the aggregated feature representations to the target GS image in a robust way.
Our symmetric undistortion network (SUNet) can be trained end-to-end and solely uses the ground truth GS image for supervision.
Furthermore, it can also inpaint the occluded regions from the learned image priors, resulting in a high-quality GS image as shown in Fig.~\ref{fig:overview}(g).
Experimental results on two benchmark datasets demonstrate that our approach is superior to the state-of-the-art methods in removing the RS artifacts.

In summary, our main contributions are:
\begin{itemize}
	\vspace{-1.5mm}
	\item We propose an efficient end-to-end symmetric RS undistortion network to solve the generic RS correction problem with two consecutive frames.
	\vspace{-1.5mm}
	\item Our context-aware cost volume together with the symmetric consistency constraint can aggregate the contextual cues of two input RS images effectively.
	\vspace{-1.5mm}
	\item Extensive experiments show that our approach performs favorably against the state-of-the-art methods in both GS image restoration and inference efficiency.
\end{itemize}

\vspace{-3.5mm}
\section{Related Work} \label{sec:RelatedWork}
\vspace{-1.5mm}
We categorize the relevant works into classical model based and deep learning based RS correction methods. 
They can also be further subdivided into single-frame-based and multi-frame-based methods, respectively.

\vspace{1mm}
\noindent\textbf{Classical model based RS correction methods.} 
Many techniques have been implemented for RS image correction from two or more RS frames.
The RS correction was posed as a temporal super-resolution problem in \cite{baker2010removing} to mitigate RS wobble in an RS video stream.
\cite{grundmann2012calibration} employed a homography mixture to achieve joint RS removal and video stabilization.
\cite{ringaby2012efficient,forssen2010rectifying} assumed that the RS camera has either pure rotation or in-plane translational motion.
An RS-aware warping \cite{zhuang2017rolling} was proposed to rectify RS images based on a differential formulation, where linear solvers were developed to recover the relative pose of the RS camera that experienced a specific motion between two consecutive frames.
They further refined both the camera motion and the depth map from dense correspondences to perform RS correction, which also determined that it relied too much on optical flow.
The occlusion-aware undistortion method \cite{vasu2018occlusion} removed the depth-dependent RS distortions from a specific setting of $\ge3$ RS images, assuming a piece-wise planar 3D scene.
Such methods rely on computationally expensive non-linear optimizations by inputting a large number of correspondences.
More recently, \cite{zhuang2020homography} proposed a differential RS homography model together with a minimal solver to account for the underlying scanline-varying poses of RS cameras, which can be used to perform RS-aware image stitching and rectification at one stroke.
\cite{albl2020two} explored a simple two-camera rig, mounted to have different RS directions, to undistort the RS images acquired by a smartphone.

Removing RS artifacts based on a single image frame is inherently a highly ill-posed task. To make it tractable, some external constraints about camera motion or scene structure need to be enforced.
\cite{rengarajan2016bows} attempted to exploit the presence of straight lines in urban scenes.
\cite{Purkait2018Minimal} modeled the Ackermann motion based on a known vertical direction \cite{zhao2021homography}, while \cite{purkait2017rolling} leveraged a prior to the Manhattan world.
\cite{lao2018robust} relaxed the Manhattan assumption and required merely the (curved) images of straight 3D lines to undistort a single RS image.
\cite{rengarajan2016bows,lao2018robust,purkait2017rolling} also assumed that the camera underwent the pure rotational motion.
Hence, they cannot work well if these underlying assumptions on scene structures and camera motions do not hold.

\vspace{1mm}
\noindent\textbf{Deep learning based RS correction methods.}
The success of deep learning in high-level vision tasks has been gradually extended to the RS geometry estimation problem (camera motion and scene structure), where a convolutional neural network (CNN) was trained to warp the RS images to their perspective GS counterparts.
Rengarajan et al. \cite{rengarajan2017unrolling} proposed the first CNN to correct a single RS image by assuming a simple affine motion model.
Afterward, Zhuang et al. \cite{zhuang2019learning} extended \cite{rengarajan2017unrolling} to learn the underlying scene structure and camera motion from a single RS image, followed by a post-processing step to produce a geometrically consistent GS image.	
Note that they often follow a general rule popular in classical methods (\emph{e.g.}, \cite{zhuang2017rolling,zhuang2020homography,lao2018robust,purkait2017rolling,rengarajan2017unrolling,wang2020relative}) that returns the GS image under the first scanline time ($0$ or $\tau$).
Very recently, Liu et al. \cite{liu2020deep} used two consecutive RS images as input and designed a deep shutter unrolling network to predict the GS image corresponding to the middle time of the second frame ($\frac{3\tau}{2}$).
To the best of our knowledge, our symmetric RS undistortion network is the first that is developed to learn the mapping from two input consecutive RS frames to the target GS frame corresponding to the intermediate time of these two frames ($\tau$).

\begin{figure*}[!t]
	\centering
	\includegraphics[width=0.99\textwidth]{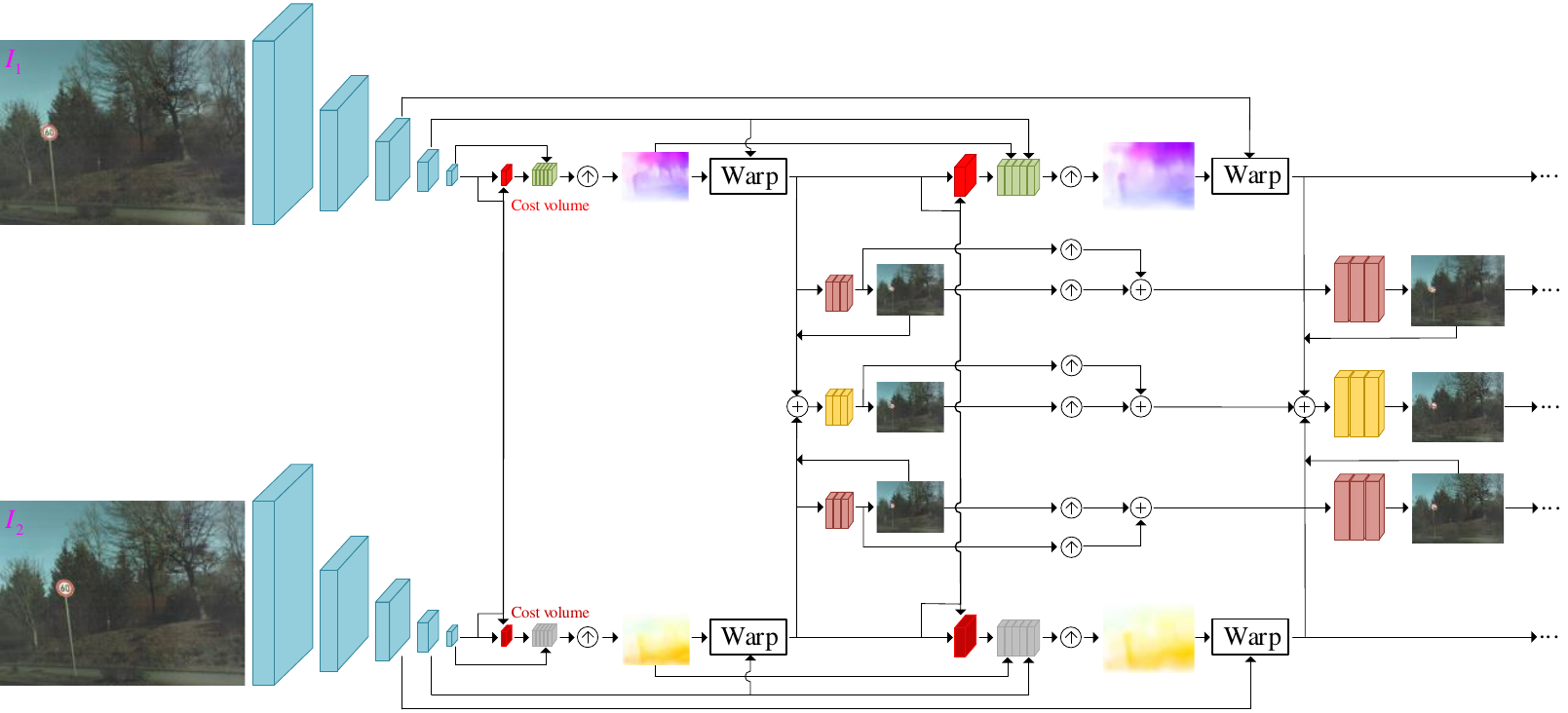}
	\caption{Overall network architecture. It mainly consists of two sub-networks: a PWC-based undistortion flow estimator and a time-centered GS image decoder. We only show the RS correction modules at the top two levels. For the rest of the pyramidal levels (excluding the first two layers), the overall RS correction modules have a similar structure as the second to the top level. Note that only the second to fifth pyramid features are warped, following a tailored correlation GS image decoder. Our network is designed symmetrically to aggregate two consecutive RS images in a coarse-to-fine manner. The symmetric convolutional layers of the same color share the same weights.\label{fig:architecture}}
	\vspace{-3.8mm}
\end{figure*}

\section{Approach} \label{sec:Approach}
Our network accepts two consecutive RS images and outputs a corrected GS image that corresponds to the intermediate time of these two frames.
To effectively exploit and aggregate contextual information of two consecutive RS images, our pipeline is symmetrically constructed and consists of two main parts, \emph{i.e.}, a PWC-based undistortion flow estimator network and a time-centered GS image decoder network, as shown in Fig.~\ref{fig:architecture}.
Henceforth, let ${\bm I}$ denote the image, ${\bm c}$ the feature representation, and ${\bm F}$ the undistortion flow. Meanwhile, in the subscript, ${t \in \{1,2\}}$ indicates the $t$-th RS image, ${t \to g}$ shows the corresponding GS counterpart warped from the $t$-th RS content, and $g$ corresponds to the time-centered corrected GS instance. Particularly, the superscript ${l}$ represents the products of $\frac{1}{2^{l-1}}$ resolution corresponding to the $l$-th pyramid level. 
Since our network is symmetric, next we will describe only the network architecture associated with the first RS image.

We first build a weight-sharing feature pyramid for two input RS images ${\bm I}_1$ and ${\bm I}_2$.
At the top $L$-th pyramid level, we construct a cost volume by comparing features of a pixel in the first RS image with corresponding features in the second RS image. As the top level is of small spatial resolution, we can construct the cost volume using a small search range.
The cost volume and features ${\bm c}_{1}^{L}$ of the first image are then fed to a CNN to estimate the undistortion flow ${\bm F}_{1 \to g}^{L-1}$ followed by an upsampling operation.
The upsampled undistortion flow ${\bm F}_{1 \to g}^{L-1}$ is then delivered to the next pyramid level.
At the second to the top level, we warp features ${\bm c}_{1}^{L-1}$ of the first RS image to its GS counterpart ${\bm c}_{1 \to g}^{L-1}$ using this upsampled undistortion flow.
Meanwhile, ${\bm c}_{1 \to g}^{L-1}$ is fed to a CNN to predict the forward GS image ${\bm I}_{1 \to g}^{L-1}$. The second RS image is similarly processed. 
Further, ${\bm I}_{1 \to g}^{L-1}$, ${\bm I}_{2 \to g}^{L-1}$, ${\bm c}_{1 \to g}^{L-1}$, and ${\bm c}_{2 \to g}^{L-1}$ are concatenated into ${\bm c}_{g}^{L-1}$, which is then decoded by CNN layers to recover the time-centered corrected GS image ${\bm I}_{g}^{L-1}$.
Next, we construct a context-aware cost volume using features ${\bm c}_{1 \to g}^{L-1}$ and ${\bm c}_{2 \to g}^{L-1}$ warped from the first and the second RS images respectively.
As warping compensates for the large motion, we can still use a small search range to construct the cost volume.
This cost volume, features ${\bm c}_{1}^{L-1}$ of the first RS image, and the upsampled undistortion flow ${\bm F}_{1 \to g}^{L-1}$ are fed to a CNN to estimate a square-scale undistortion flow ${\bm F}_{1 \to g}^{L-2}$ at the $(L-1)$-th level. These processes repeat until the desired level.

In the following, we introduce the key components of each module, including pyramid feature extractor, undistortion flow estimator, and time-centered GS image decoder networks. 
The details of the architecture are provided in Appendix B.

\vspace{1mm}
\noindent\textbf{Feature pyramid extractor.}
For two consecutive RS images ${\bm I}_1$ and ${\bm I}_2$, we encode the $L$-level pyramids of feature representations to explore richer multi-scale information. Note that the bottom ($0$-th) level is the input images, \emph{i.e.}, ${\bm c}_{t}^{0}={\bm I}_{t}$. 
We utilize a single 2D convolution with a stride of 2 to downsample the features ${\bm c}_{t}^{l-1}$ at the $(l-1)$-th pyramid level, and obtain the feature representation ${\bm c}_{t}^{l}$ at the $l$-th layer. We do not downsample the $0$-th pyramid, \emph{i.e.}, ${\bm c}_{t}^{0}$ and ${\bm c}_{t}^{1}$ have the same size. Note that the $1$-th pyramid is used as a transitional layer.
Specifically, three ResNet blocks \cite{he2016deep} are performed after each downsampling operation.
In our implementation, we use a $6$-level pyramid, \emph{i.e.}, $L=5$, consisting of 5 levels of CNN features and the input RS images as the bottom level. 
Wherein, the number of feature channels is set to 16, 32, 64, 96, and 128, respectively.

\vspace{1mm}
\noindent\textbf{Context-aware cost volume layer.}
After obtaining the upsampled undistortion flow ${\bm F}_{t \to g}^{l-1}$ at the $l$-th layer, we warp the feature ${\bm c}_{t}^{l-1}$ of the $t$-th RS image to its GS counterpart ${\bm c}_{t \to g}^{l-1}$ using the forward warping block \cite{liu2020deep}, which can compensate for RS distortions and put the corrected GS image patches at the right scale.
The warped features are the same as the pyramid features at the top level, \emph{i.e.}, ${\bm c}_{t \to g}^{L}={\bm c}_{t}^{L}$.
We further construct a cost volume ${\bm {cv}}_{1 \to 2}^{l-1}$ as the correlation  \cite{dosovitskiy2015flownet,sun2018pwc} between ${\bm c}_{1 \to g}^{l-1}$ and ${\bm c}_{2 \to g}^{l-1}$ by taking ${\bm c}_{2 \to g}^{l-1}$ as a reference. Similarly, with ${\bm c}_{1 \to g}^{l-1}$ as a reference, ${\bm {cv}}_{2 \to 1}^{l-1}$ can also be obtained accordingly.
A search range of $d$ pixels is used to compute the cost volume at each level.
Note that this is more reasonable than the ill-aligned embedding of cost volume in \cite{liu2020deep}.
Building these multi-resolution matching costs can promote mutual consistency between the contextual warped feature representations ${\bm c}_{1 \to g}^{l-1}$ and ${\bm c}_{2 \to g}^{l-1}$, which is conducive to improving the fidelity of the subsequent corrected GS images decoded by the aggregated features.

\vspace{1mm}
\noindent\textbf{Undistortion flow estimator.} 
Next, we feed features ${\bm c}_{t}^{l-1}$ of the $t$-th RS image, the undistortion flow ${\bm F}_{t \to g}^{l-1}$, and the corrsponding cost volume into five DenseNet blocks \cite{huang2017densely}. The undistortion flow ${\bm F}_{t \to g}^{l-2}$ is then estimated followed by an upsampling operation at the $(l-1)$-th level.
To connect the subsequent time-centered GS image decoder modules, we set the desired level to be $l_0$, \emph{i.e.}, our model outputs multi-scale undistortion flows ${\bm F}_{t \to g}^{l}$, $L-1 \ge l \ge l_0-1$, which can be used to estimate the multi-scale GS images with different resolutions.
Note that the number of feature channels for these five DenseNet blocks is respectively 128, 128, 96, 64, and 32 at the top pyramid level. Decreasingly, we have 64, 64, 48, 32, 16 for the $(L-1)$-th pyramid level and 32, 32, 24, 16, 8 for the $(L-2)$-th pyramid level.

\vspace{1mm}
\noindent\textbf{Time-centered GS image decoder.}
Inspired by the image decoder proposed in \cite{liu2020deep}, we employ three ResNet blocks \cite{he2016deep}, whose structure is consistent with that in the feature pyramid extractor, followed by a GS image prediction layer and a deconvolutional layer.
The warped features ${\bm c}_{1 \to g}^{l-1}$ of the first RS image, the warped features ${\bm c}_{2 \to g}^{l-1}$ of the second RS image, and the concatenated features ${\bm c}_{g}^{l-1}$ are embedded to generate forward and backward GS images ${\bm I}_{1 \to g}^{l-1}$ and ${\bm I}_{2 \to g}^{l-1}$, and a target GS image ${\bm I}_{g}^{l-1}$ between $l$-th and $(l\!-\!1)$-th pyramid levels, respectively.
Note that the concatenated features ${\bm c}_{g}^{l-1}$ fuse ${\bm c}_{t \to g}^{l-1}$, ${\bm I}_{t \to g}^{l-1}$, and features of the previous level (if it exists) in an adaptive selection manner.
Our network finally outputs a half-resolution GS image and we use bilinear upsampling followed by a 2D convolution to obtain the full-resolution time-centered GS image. 

\vspace{1mm}
\noindent\textbf{Training loss.}
Given a pair of consecutive RS images $\{{\bm I}_t\}_1^2$, our SUNet predicts the undistortion flows ${\bm F}_{t \to g}^{l-1}$, the forward/backward GS images ${\bm I}_{t \to g}^{l-1}$, and the corrected GS image ${\bm I}_{g}^{l-1}$ at the $l$-th pyramid level ($l \ge l_0$).
Let ${\bm I}_{GT}^{l-1}$ denote the corresponding ground truth (GT) GS image in the intermediate time of two consecutive RS frames.
Note that the superscript $^{l-1}$ indicates $\frac{1}{2^{l-2}}$ resolution images.
Our loss function can be formulated as:
\begin{equation}\label{eq:1}
\small
\mathcal{L} = \lambda_r\mathcal{L}_r + \lambda_p\mathcal{L}_p + \lambda_c\mathcal{L}_c + \lambda_s\mathcal{L}_s.
\end{equation}

\emph{Reconstruction loss $\mathcal{L}_r$.} We model the pixel-wise reconstruction quality of the corrected GS image on multiple scales as:
\begin{equation}\label{eq:2}
\small
\mathcal{L}_{r} = \sum_{l=l_0-1}^{L}\left\|{\bm I}_{GT}^{l-1}-{\bm I}_{g}^{l-1}\right\|_{1},
\end{equation}
where all images are defined in the RGB space.

\emph{Perceptual loss $\mathcal{L}_p$.} To mitigate the blurry effect in the corrected GS image, similar to \cite{liu2020deep}, we employ a perceptual loss $\mathcal{L}_p$ \cite{johnson2016perceptual} to preserve details of the predictions and make estimated GS image sharper. $\mathcal{L}_p$ is defined as:
\begin{equation}\label{eq:3}
\small
\mathcal{L}_{p} = \sum_{l=l_0-1}^{L}\left\| \phi \left( {\bm I}_{GT}^{l-1} \right) - \phi \left( {\bm I}_{g}^{l-1} \right) \right\|_{1},
\end{equation}
where $\phi$ represents the $conv3\_3$ feature extractor of the VGG19 model \cite{simonyan2014very}.

\emph{Consistency loss $\mathcal{L}_c$.} To combine cues from ${\bm I}_1$ and ${\bm I}_2$, we enforce their respective warped feature representations ${\bm c}_{1 \to g}^{l-1}$ and ${\bm c}_{2 \to g}^{l-1}$ to be as close to each other as possible.
This operation also facilitates subsequent alignment in the concatenated feature representation ${\bm c}_{g}^{l-1}$, thereby alleviating artifacts in the corrected GS image ${\bm I}_{g}^{l-1}$ predicted by ${\bm c}_{g}^{l-1}$.
Specifically, we introduce the pixel-wise consistency loss $\mathcal{L}_c$ to supervise the network to align the forward and backward images ${\bm I}_{1 \to g}^{l-1}$ and ${\bm I}_{2 \to g}^{l-1}$ predicted by the first and the second RS images respectively across different levels, so that it can recover more details. 
$\mathcal{L}_c$ is defined as:
\begin{equation}\label{eq:4}
\small
\mathcal{L}_{c} = \sum_{t=1}^{2}\sum_{l=l_0}^{L}\left\|{\bm I}_{GT}^{l-1}-{\bm I}_{t \to g}^{l-1}\right\|_{1}.
\end{equation}

\emph{Smoothness loss $\mathcal{L}_s$.} Finally, we add a smoothness term \cite{liu2017video} to encourage piecewise smoothness in the estimated undistortion flows as:
\begin{equation}\label{eq:5}
\small
\mathcal{L}_{s} = \sum_{t=1}^{2}\sum_{l=l_0}^{L}\left\|\nabla {\bm F}_{t \to g}^{l-1}\right\|_{2}.
\end{equation}

Note that the context-aware cost volume and warping layers have no learnable parameters.
Our network can be end-to-end trained as each of its modules is differentiable.

\section{Experimental Results} \label{sec:Experiments}
\noindent\textbf{Datasets.}
As far as we know, the two RS datasets Carla-RS and Fastec-RS published in \cite{liu2020deep} are the only RS datasets that provide ground truth GS supervisory signals and are therefore suitable for our task.
The Carla-RS dataset is generated from a virtual 3D environment using the Carla simulator \cite{dosovitskiy2017carla}, involving general six degrees of freedom motions.
The Fastec-RS dataset contains real-world RS images synthesized by a professional high-speed GS camera.
Following the evaluation in \cite{liu2020deep}, we also test our algorithm on these two benchmarks.
Similarly, the Carla-RS dataset is divided into a training set of 210 sequences and a test set of 40 sequences, and the Fastec-RS dataset has 56 sequences for training and 20 sequences for the test.
There are no overlapping scenarios between the training set and the test set.
Note that the Carla-RS dataset provides the ground truth occlusion masks.
Following \cite{liu2020deep}, we set quantitative evaluation experiments as the Carla-RS dataset with occlusion mask (CRM), the Carla-RS dataset without occlusion mask (CR), and the Fastec-RS dataset (FR), respectively.

\vspace{1mm}
\noindent\textbf{Implementation details.}
The network is implemented in PyTorch. 
The weights have been determined empirically as $\lambda_r=10$, $\lambda_p=1$, $\lambda_c=5$, and $\lambda_s=0.1$.
We downsample the ground truth GS images to obtain the supervision signals at different levels.
The desired level $l_0$ is set to 3 and the top pyramid layer $L$ is set to 5,
\emph{i.e.}, we only work with the feature representations captured by the second to fifth pyramid layers to estimate the undistortion flow and correct the RS images.
We set a search range of 4 pixels to compute the cost volume at each level, \emph{i.e.}, $d=4$.
We adopt the Adam optimizer \cite{Kingma_Adam_ICLR_2015} with a learning rate of $10^{-4}$. 
The chosen batch size is 6 and the network is trained for 400 epochs. 
We use a uniform random crop at a horizontal resolution of 256 pixels for data augmentation. Note that we do not change the longitudinal resolution to maintain the inherent characteristics of consecutive RS images, which can help to better learn to accumulate contextual information of two consecutive RS images.

\begin{table}[!t]
	\footnotesize
	\caption{\small Ablation study on the context-aware cost volume. Removing the cost volume (0) leads to consistently worse performances. Our network can handle large RS distortions using a small search range to compute the cost volume.}\label{t1}
	\centering
	\begin{tabular}{lcccccc}
		\hline
		\multirow{2}{*}{Max. Disp.} & \multicolumn{3}{c}{PSNR$\uparrow$}                      &   & \multicolumn{2}{c}{SSIM$\uparrow$}      \\ \cline{2-4} \cline{6-7}
		& CRM            & CR             & FR            & & CR            & FR            \\ \hline
		0                  & 21.87          & 21.84          & 25.21          & & 0.66          & 0.76          \\
		2                  & 28.76          & 28.60          & 28.07          & & 0.84          & 0.83          \\
		4 (Ours)           & \textbf{29.28} & \textbf{29.18} & \textbf{28.34} & & \textbf{0.85} & \textbf{0.84} \\
		6                  & 29.22          & 29.11          & 28.14          & & \textbf{0.85} & 0.83          \\ \hline
	\end{tabular}
\end{table}

\begin{table}[!t]
	\footnotesize
	\caption{\small Ablation study on the size of the convolving kernel at the $1$-th pyramid level. Note that ``$0 \times 0$" means that we remove the first layer of the current feature pyramid extractor, \emph{i.e.}, $L=4$ and the new ${\bm c}_{t}^{1}$ is half the size of ${\bm c}_{t}^{0}$. The transitional feature representation across the $1$-th pyramid level can promote the better perception of large RS distortions, thereby producing substantially better results.}\label{t2}
	\centering
	\begin{tabular}{lcccccc}
		\hline
		\multirow{2}{*}{Kernel Size} & \multicolumn{3}{c}{PSNR$\uparrow$}                      &   & \multicolumn{2}{c}{SSIM$\uparrow$}      \\ \cline{2-4} \cline{6-7}
		& CRM            & CR             & FR             & & CR            & FR            \\ \hline
		$0 \times 0$       & 29.02          & 28.91          & 27.83          & & 0.84 & 0.82          \\
		$3 \times 3$       & 29.08 &  28.93 & 28.13          & & \textbf{0.85} & 0.83          \\			
		$5 \times 5$       & 29.26          & 29.16          & 28.17          & & \textbf{0.85} & 0.83          \\ 
		$7 \times 7$ (Ours) & \textbf{29.28} & \textbf{29.18} & \textbf{28.34} & & \textbf{0.85} & \textbf{0.84} \\ \hline
	\end{tabular}
\end{table}

\vspace{1mm}
\noindent\textbf{Competing methods.}
To evaluate the performance of the proposed approach, we compare with two representative two-frame-based RS correction methods \cite{zhuang2017rolling,liu2020deep} that are the most relevant baselines to our approach.
In addition, we also compare with the state-of-the-art single-frame-based RS correction method \cite{zhuang2019learning}.
Since \cite{zhuang2019learning} does not release their code, we follow the reimplementation by \cite{liu2020deep}.
Note that the Fastec-RS dataset does not provide ground truth depth and motion, so we are unable to complete the RS correction using \cite{zhuang2019learning}.
A constant velocity model is assumed and solved to achieve RS correction in \cite{zhuang2017rolling}.
Because the deep shutter unrolling network (DSUN) \cite{liu2020deep} recover the GS image corresponding to the time of the middle scanline of the second RS frame, we thus adjust and retrain it to predict the GS image corresponding to the intermediate time of two consecutive RS frames for consistent and fair comparisons.

\vspace{1mm}
\noindent\textbf{Evaluation metrics.}
Following previous works, we use PSNR and SSIM metrics to report the quantitative results of our method.
The larger the PSNR/SSIM score, the higher the quality of the corrected GS image.

\begin{table}[!t]
	\footnotesize
	\caption{\small Ablation study on whether to multiply the time offsets. Compared with \cite{liu2020deep}, our symmetric network can implicitly learn the dependence of RS undistortion flow on scanline time.}\label{t3}
	\centering
	\begin{tabular}{lcccccc}
		\hline
		\multirow{2}{*}{Time Offset} & \multicolumn{3}{c}{PSNR$\uparrow$}                      &   & \multicolumn{2}{c}{SSIM$\uparrow$}      \\ \cline{2-4} \cline{6-7}
		& CRM            & CR             & FR             & & CR             & FR        \\ \hline
		Yes       & 28.56          & 28.45          & 28.13          & & 0.83          & 0.83     \\
		No (Ours) & \textbf{29.28} & \textbf{29.18} & \textbf{28.34} & & \textbf{0.85} & \textbf{0.84} \\ \hline
	\end{tabular}
\end{table}

\begin{table}[!t]
	\footnotesize
	\caption{\small Ablation study on the consistency loss. $\lambda_c=0$ means no consistency loss is used. The self-supervised consistency loss is defined as measuring only the difference between forward and backward GS images. Our loss function is effective to align contextual cues, especially the Fastec-RS dataset.}\label{t4}
	\centering
	\begin{tabular}{lcccccc}
		\hline
		\multirow{2}{*}{Consist. Loss} & \multicolumn{3}{c}{PSNR$\uparrow$}                      &   & \multicolumn{2}{c}{SSIM$\uparrow$}      \\ \cline{2-4} \cline{6-7}
		& CRM            & CR           & FR             & & CR            & FR            \\ \hline
		$\lambda_c=0$       & 29.05     & 28.94          & 27.89          & & 0.84 & 0.82          \\
		Self-supervised     & 29.15     & 28.99          & 28.02          & & \textbf{0.85} & 0.83          \\			
		Ours  & \textbf{29.28} & \textbf{29.18} & \textbf{28.34} & & \textbf{0.85} & \textbf{0.84} \\ \hline
	\end{tabular}
\end{table}

\begin{table}[!t]
	\footnotesize
	\caption{\small Effectiveness of different combinations of training losses.}\label{s_t2}
	\centering
	\begin{tabular}{lcccccc}
		\hline
		\multirow{2}{*}{} & \multicolumn{3}{c}{PSNR$\uparrow$}                      &   & \multicolumn{2}{c}{SSIM$\uparrow$}      \\ \cline{2-4} \cline{6-7}
		& CRM            & CR             & FR            & & CR            & FR            \\ \hline
		w/o $\mathcal{L}_r$          & 28.00          & 27.90          & 27.29          & & 0.83    & 0.81          \\
		w/o $\mathcal{L}_p$          & 29.08          & 28.95          & 28.20          & & \textbf{0.85}    & \textbf{0.84}          \\
		w/o $\mathcal{L}_c$          & 29.05          & 28.94          & 27.89          & & 0.84    & 0.82          \\
		w/o $\mathcal{L}_s$     & 29.19          & 28.07          & 28.15          & & \textbf{0.85}    & 0.83          \\\hline
		full loss    & \textbf{29.28} & \textbf{29.18} & \textbf{28.34} & & \textbf{0.85} & \textbf{0.84}  \\\hline
	\end{tabular}
\end{table}

\vspace{1mm}
\noindent\textbf{Ablation studies.} We study the role of each proposed component in SUNet, for further insight into the design choices.

\emph{\textbf{1) Context-aware cost volume.}}
We analyze the effect of different search range in the cost volume on the performance of our approach, shown in Table \ref{t1}.
One can see that removing the cost volume is consistently bad for GS image restoration, which indicates that the cost volume can effectively promote the perception and alignment of the warped contexts of two consecutive RS images.
Thus the contextual cost volume can be regarded as a core unit of our network to deal with the non-local operations of the RS correction.
Furthermore, a 4-pixel search range is enough to undistort the RS images successfully, which is due to the fact that a 4-pixel search range is capable of handling up to 100-pixel displacements at the input resolution.
Specifically, a larger range leads to similar performance, so a larger search range may bring better benefits when the image resolution increases or the RS distortion is severe.

\begin{figure*}[!t]
	\centering
	\includegraphics[width=0.98\textwidth]{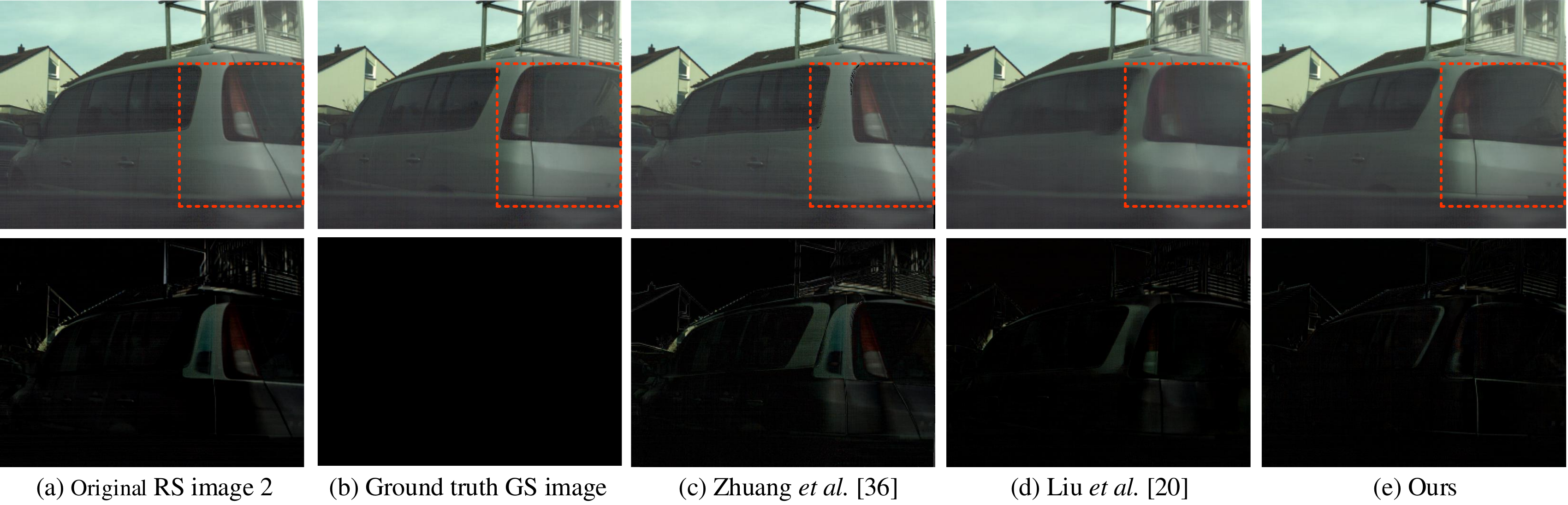}
	\caption{Qualitative results against baseline methods on the Fastec-RS dataset. Even rows: absolute difference between the corresponding image and the ground truth GS image. (c-e) GS images predicted by Zhuang \emph{et al.} \cite{zhuang2017rolling}, Liu \emph{et al.} \cite{liu2020deep}, and our approach, respectively.\label{fig:exp_Fastec_1}}
\end{figure*}

\emph{\textbf{2) Kernel size.}}
Although we do not directly utilize the feature representation of the first pyramid level, we investigate its influence as a transitional layer on the final performance, shown in Table \ref{t2}.
First, we directly remove the first layer of the current feature pyramid extractor, \emph{i.e.}, a 5-level pyramid is used.	
Then we test different convolution kernel sizes at the first pyramid level.
All these settings have similar performance in terms of PSNR and SSIM, but we empirically observe that a $7 \times 7$ kernel size has relatively better correction capabilities on the foreground, resulting in better visual quality.
One possible explanation is that the foreground objects exist more serious RS distortion, and a larger convolution kernel can increase the receptive field, which therefore contributes to recovering more geometrically consistent GS images.

\emph{\textbf{3) Time offset.}}
The undistortion flow depends not only on the camera motion and the 3D scene geometry but also on the scanline time of a particular pixel. 
Thus, \cite{liu2020deep} explicitly multiplies the velocity field with the time offset to model the adapted undistortion flow.
Similarly, we also explicitly multiply our upsampled undistortion flows by the corresponding normalized time offsets between the exposure time of the captured pixel and the intermediate time of the two frames.
Interestingly, the results in Table \ref{t3} demonstrate that our SUNet can directly return the suitable RS-aware undistortion flows to transform the learned feature representations to their corresponding GS counterparts.
We reckon this is because explicit modeling of time offset at multi-resolution may confuse the network, while our PWC-based symmetric architecture can successfully learn the inherent scanline-dependent characteristics of the undistortion flow.
See Appendix A for more analyses.

\emph{\textbf{4) Loss function.}}
We also perform two experiments to prove the effectiveness of the proposed consistency loss $\mathcal{L}_c$. One is to remove $\mathcal{L}_c$ from the loss function $\mathcal{L}$ in Eq.~\ref{eq:1}, \emph{i.e.}, $\lambda_c=0$. The other is to change the consistency metric in Eq.~\ref{eq:4} to the direct difference between forward/backward GS images ${\bm I}_{1 \to g}^{l-1}$ and ${\bm I}_{2 \to g}^{l-1}$ without introducing the ground truth GS images ${\bm I}_{GT}^{l-1}$, where $\lambda_c$ is still fixed at 5, which constitutes a self-supervised metric.
As can be seen from Table~\ref{t4}, removing the consistency loss consistently weakens the performance of GS image restoration, and the self-supervised metric is not effective enough, especially in the Fastec-RS dataset. 
In addition, they will cause visible seams in the foreground, such as the signboard in Fig.~\ref{fig:overview}.
We also find that they all degrade the estimates of both ${\bm I}_{1 \to g}^{l-1}$ and ${\bm I}_{2 \to g}^{l-1}$ to black, which sacrifices the interpretability of the network in the geometric sense.
One possible reason is that the occlusion between ${\bm I}_{1 \to g}^{l-1}$ and ${\bm I}_{2 \to g}^{l-1}$ is more serious than that between them and ${\bm I}_{GT}^{l-1}$, which degrades the predictions of both ${\bm I}_{1 \to g}^{l-1}$ and ${\bm I}_{2 \to g}^{l-1}$ and thus contributes little to the alignment of the contextual cues.
Overall, it proves our loss function in Eq.~\ref{eq:1} can fully supervise the RS correction.

\emph{\textbf{5) Training loss.}}
We report the impact of different combinations of the loss terms in training our model, as shown in Table~\ref{s_t2}. Note that removing the perceptual loss $\mathcal{L}_p$ will cause the corrected image to appear blurred effect, and the reconstruction loss $\mathcal{L}_r$ is particularly important. Our total loss function yields the best model, which facilitates better removal of RS artifacts to produce high-quality results.

\begin{table}[!t]
	\footnotesize
	\caption{\small Quantitative comparisons of the performance between our approach and the state-of-the-art baseline methods. Note that we cannot use \cite{zhuang2019learning} to benchmark the Fastec-RS dataset due to its lack of training ground truth. Our approach again performs the best.}\label{t5}
	\centering
	\begin{tabular}{lcccccc}
		\hline
		\multirow{2}{*}{Methods} & \multicolumn{3}{c}{PSNR$\uparrow$ (dB)}                      &   & \multicolumn{2}{c}{SSIM$\uparrow$}      \\ \cline{2-4} \cline{6-7}
		& CRM            & CR             & FR            & & CR            & FR            \\ \hline
		Single-frame \cite{zhuang2019learning}  & 18.70          & 18.47          & -          & & 0.58          & -          \\
		Model-based \cite{zhuang2017rolling}    & 25.93          & 22.88          & 21.44      & & 0.77          & 0.71          \\
		DSUN \cite{liu2020deep}                 & 26.90          & 26.46          & 26.52      & & 0.81          & 0.79          \\
		SUNet (Ours)          & \textbf{29.28} & \textbf{29.18} & \textbf{28.34} & & \textbf{0.85} & \textbf{0.84} \\ \hline
	\end{tabular}
\end{table}

\begin{figure*}[!t]
	\centering
	\includegraphics[width=0.99\textwidth]{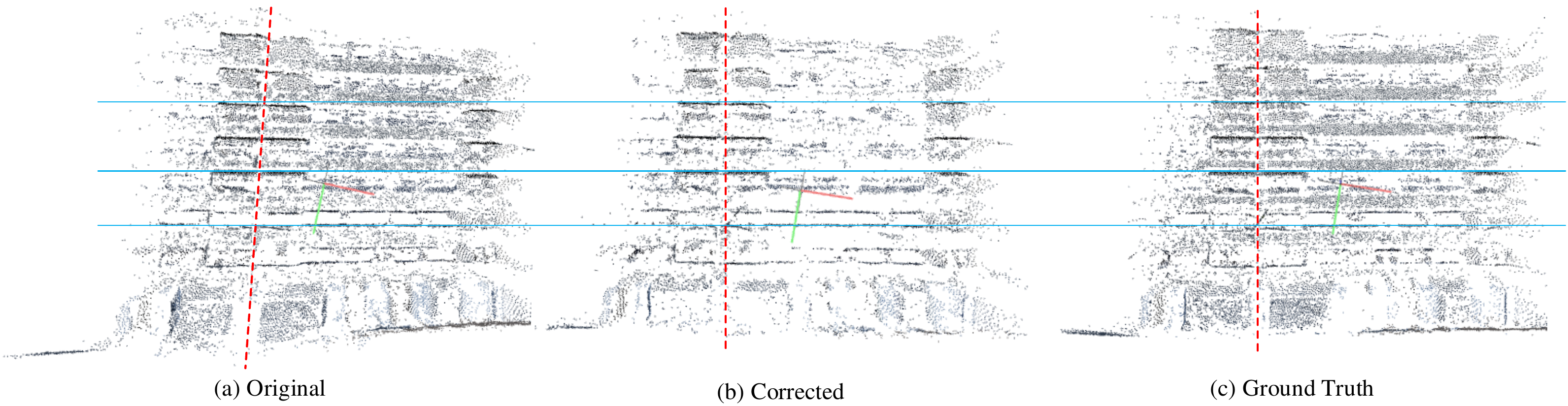}
	\caption{SfM results by Colmap \cite{schonberger2016structure} for a building. (a) Reconstructed 3D model with original RS images. (b) Reconstructed 3D model with corrected GS images. (c) Reconstructed 3D model with ground truth GS images. It demonstrates that our pipeline removes the undesired RS distortion and generates a more accurate 3D model as the ground truth 3D model.\label{fig:construction}}
\end{figure*}
\begin{figure*}[!t]
	\centering
	\includegraphics[width=0.92\textwidth]{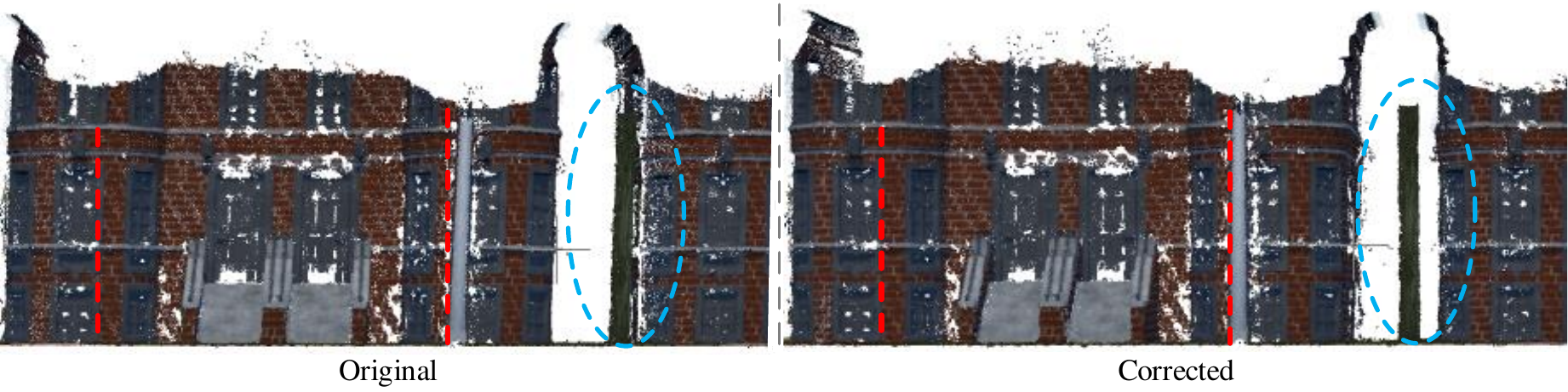}
	\caption{A 3D reconstruction example using the original RS images and our corrected GS images, respectively.\label{fig:construction_add}}
\end{figure*}

\begin{figure*}[!t]
	\centering
	\includegraphics[width=1.0\textwidth]{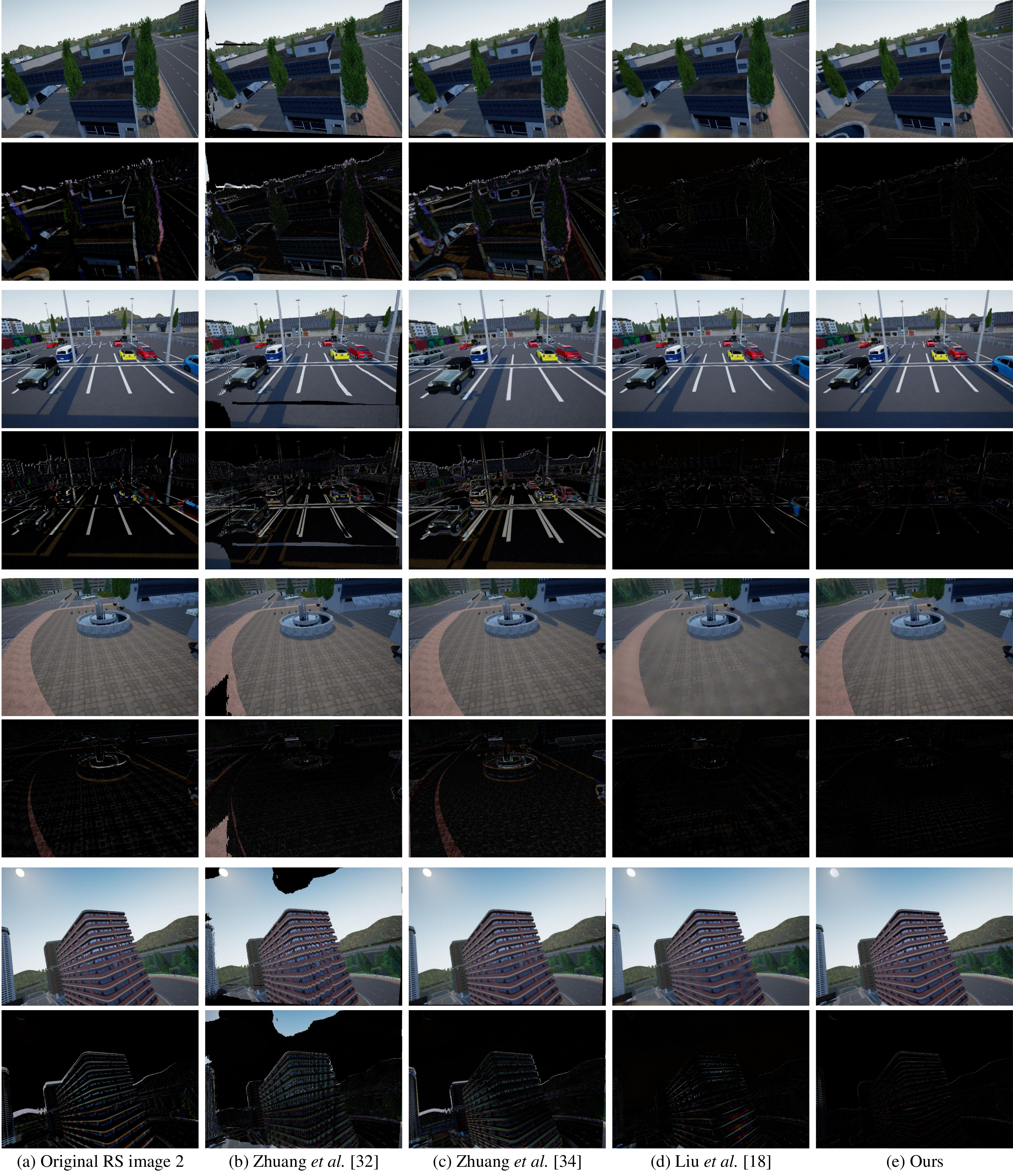}
	\caption{Qualitative results against baseline methods. Even rows: absolute difference between the corresponding image and the ground truth GS image. (a) The original second frame RS images. (b-e) GS images predicted by Zhuang \emph{et al.} \cite{zhuang2017rolling}, Zhuang \emph{et al.} \cite{zhuang2019learning}, Liu \emph{et al.} \cite{liu2020deep}, and our SUNet method, respectively.\label{fig:exp_Carla}}
\end{figure*}

\vspace{1.0mm}
\noindent\textbf{Comparison with baseline methods.}
We report the quantitative and qualitative comparisons against the baseline methods in Table \ref{t5} and Fig.~\ref{fig:exp_Fastec_1}, respectively.
We also provide more qualitative comparisons in Fig.~\ref{fig:exp_Carla}.
The experimental results demonstrate that our approach outperforms the three state-of-the-art methods by a significant margin.
Note that the deep shutter unrolling network (DSUN) \cite{liu2020deep} is most relevant to our approach. However, DSUN is difficult to restore the GS image details that have not been seen by the second RS image, which also validates our preceding analysis in Section \ref{sec:Introduction}. The visually unpleasing areas, \emph{i.e.}, the lower parts of the corrected GS image, are mainly concentrated in the scanlines with a large time offset from the intermediate time of two input frames. On the contrary, our approach can overcome these obstacles quite well, as shown in Fig.~\ref{fig:exp_Fastec_1}.
Also, the forward and backward GS images rectified from the first frame and the second frame respectively are depicted in Fig.~\ref{fig:overview}(e)\&(f), which indicates the importance of contextual cues.
More analyses are shown in Appendix A.

Furthermore, \cite{zhuang2017rolling} is a classical model based RS correction method with two consecutive frames by assuming a constant velocity motion model. However, its performance in poorly textured regions is not satisfactory.
\cite{zhuang2019learning} uses a single RS image as input to achieve RS correction, but it shows limited generalization performance on the Carla-RS dataset. 
A possible reason is that the estimation of camera motion and scene depth is degraded due to unseen scene contents in the test data. Also, the specific RS camera model assumptions in post-processing may have poor adaptability to this dataset.
In contrast, our data-driven approach solves a general two-frame-based RS correction problem without relying on specific RS camera model assumptions, thereby achieving better performances.
It is also worth mentioning that, benefit from the powerful expressive power of CNNs, our approach can effectively fill the occluded regions, which is unable to be reconstructed by \cite{zhuang2017rolling} and \cite{zhuang2019learning}.
An intuitive analysis is that on the Carla-RS dataset, the difference of PSNR in \cite{zhuang2017rolling} with or without the mask is 3.05 dB, while ours is only 0.1 dB.

\begin{figure*}[!t]
	\centering
	\includegraphics[width=1.0\textwidth]{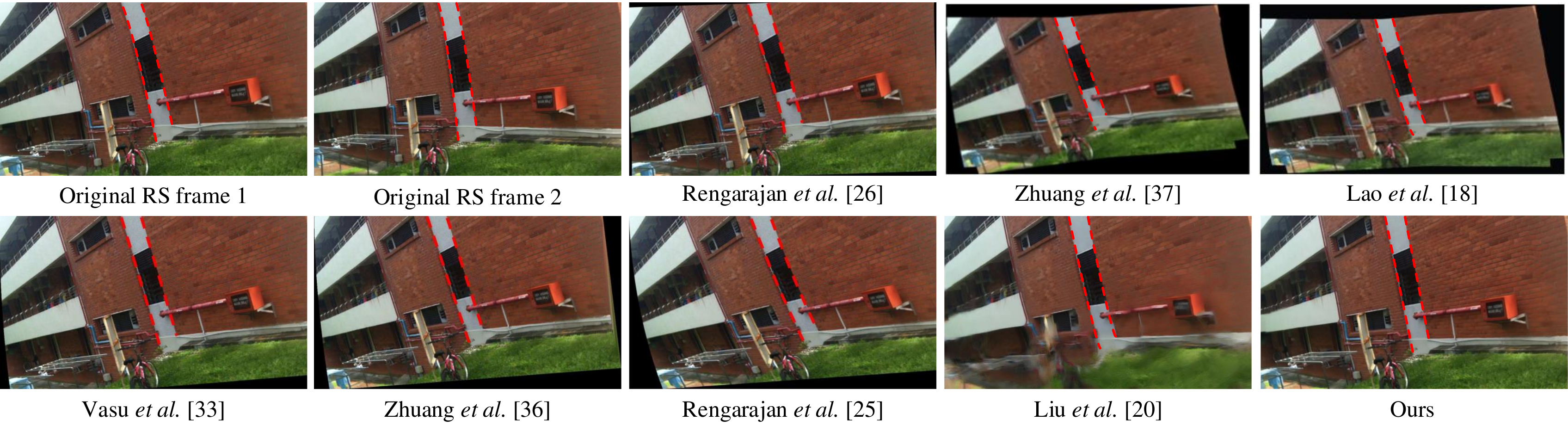}
	\caption{Qualitative comparisons against relative methods using data provided by \cite{zhuang2017rolling}. Our pipeline estimates a plausible GS image and complements the occluded regions based on the learned image prior.\label{fig:generalize}}
\end{figure*}

\vspace{1.0mm}
\noindent\textbf{3D reconstruction.}
We run an SfM pipeline (\emph{i.e.}, Colmap \cite{schonberger2016structure}) to process the original RS image sequences, the corrected GS image sequences, and the ground truth GS image sequences, respectively. Fig.~\ref{fig:construction} and Fig.~\ref{fig:construction_add} demonstrate that our approach can correct the geometric inaccuracies and generate a more accurate 3D structure that complies with the underlying ground truth 3D scene geometry. Note that one can see the obvious RS distortions (\emph{e.g.}, skew and shrink) in the 3D model obtained by the original RS images.

\begin{figure*}[!t]
	\centering
	\includegraphics[width=0.7\textwidth]{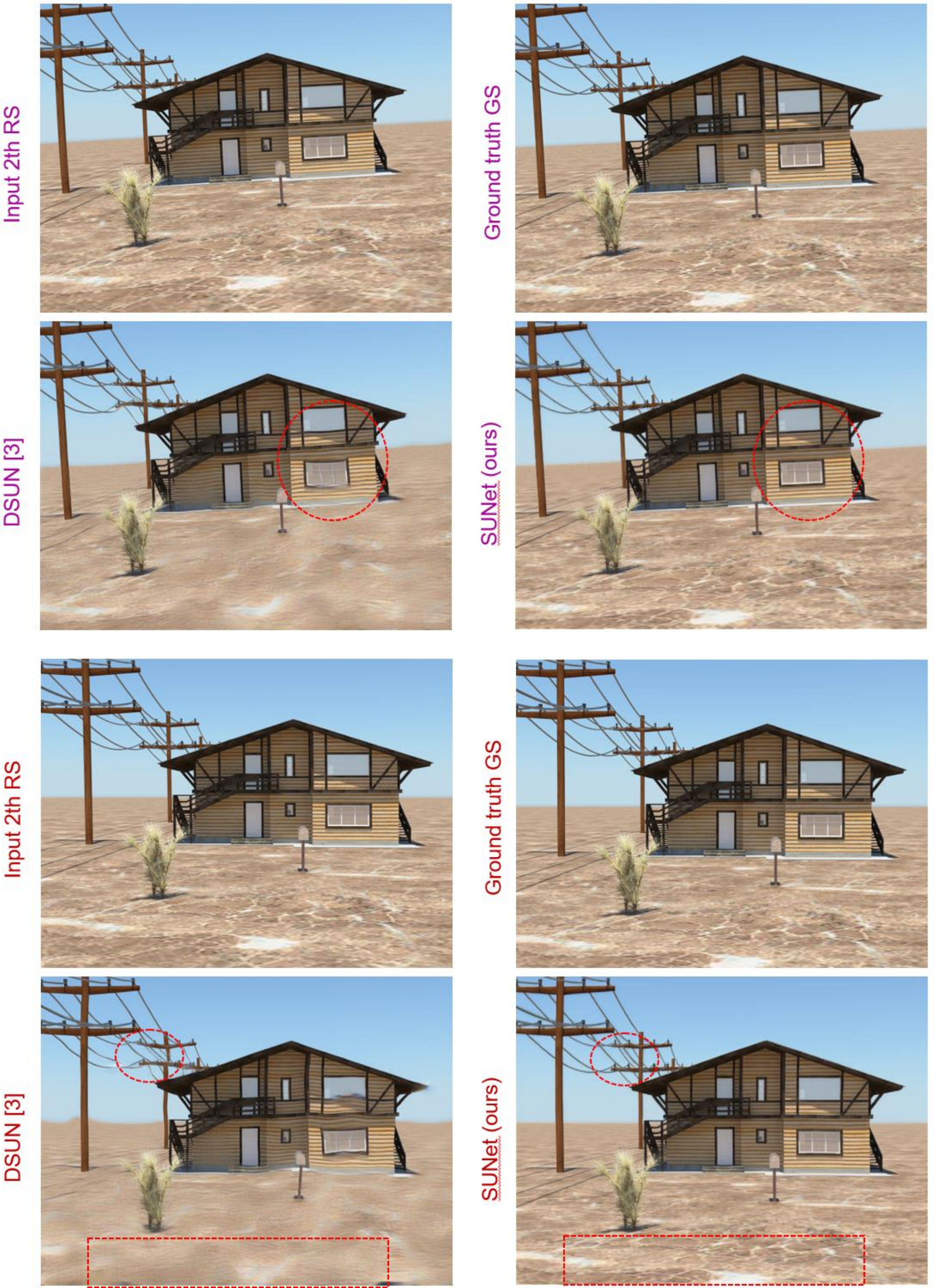}
	\caption{Two examples extracted from the demo video. DSUN \cite{liu2020deep} performs worse when dealing with unseen scenes provided by \cite{forssen2010rectifying}, while our proposed SUNet has excellent generalization performance to produce a coherent video with more detailed textures and fewer ghosting artifacts. \label{fig:video_demo}}
	\vspace{-2mm}
\end{figure*}

\begin{figure*}[!t]
	\centering
	\includegraphics[width=0.93\textwidth]{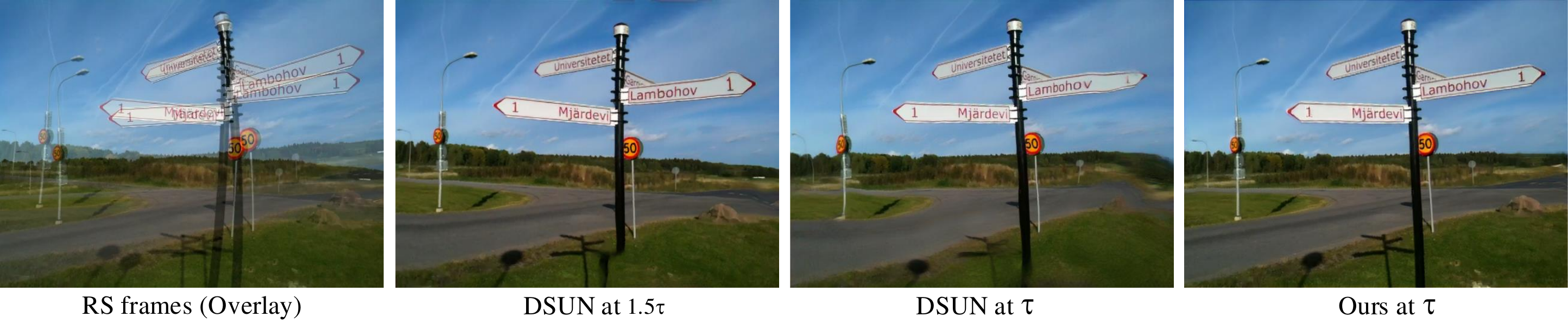}
	\caption{An example result from an RS sequence \cite{forssen2010rectifying} captured by a fast-moving iPhone 3GS camera in the real world. \label{fig:exp_real_3GS}}
\end{figure*}

\vspace{1.0mm}
\noindent\textbf{Generalization performance.}
We use the real data provided by \cite{zhuang2017rolling} to carry out the generalization experiment, and make a qualitative comparison with several relevant RS correction methods.
The results are summarized in Fig.~\ref{fig:generalize}.
To evaluate the performance of our proposed pipeline on video sequences, we further utilize the RS image dataset from \cite{forssen2010rectifying}, where each sequence consists of 12 consecutive RS frames with significant image distortions. 
Two examples are shown in Fig.~\ref{fig:video_demo}.
It is obvious that, in comparison with SUNet, DSUN \cite{liu2020deep} always fails to recover the texture on the ground.
Also, combining with Fig.~\ref{fig:exp_real_3GS}, one can further see that our approach obtains more satisfactory RS correction results with its excellent generalization ability.
Overall, our approach has good generalization ability to recover a visually compelling GS image.

\vspace{1.0mm}
\noindent\textbf{Inference time.}
Our method can correct RS images with a resolution of $640 \times 480$ pixels using an average of 0.21 seconds on an NVIDIA GeForce RTX 2080Ti GPU, which is faster than the average 0.34 seconds of DSUN \cite{liu2020deep}. 
Note that the classical two-frame-based RS correction method \cite{zhuang2017rolling} takes several minutes on an Intel Core i7-7700K CPU.

\section{Conclusion} \label{sec:Conclusion}
In this paper, we have proposed an end-to-end symmetric undistortion network for generic RS correction. Given two consecutive RS images, it can effectively estimate the GS image corresponding to the intermediate time of these two frames. Our context-aware undistortion flow estimator and the symmetric consistency enforcement can efficiently reduce the misalignment between the contexts warped from two consecutive RS images, thus achieving state-of-the-art RS correction performances. Currently, we concern ourselves with the RS correction problem corresponding to a particular time. In the future, we will explore more challenging tasks, \emph{e.g.}, a typical exposure time manipulated by the user to complete the corresponding RS correction.

\section*{Appendix A: Necessity and effectiveness of contextual aggregation}
\begin{figure*}[!t]
	\centering
	\includegraphics[width=1.0\textwidth]{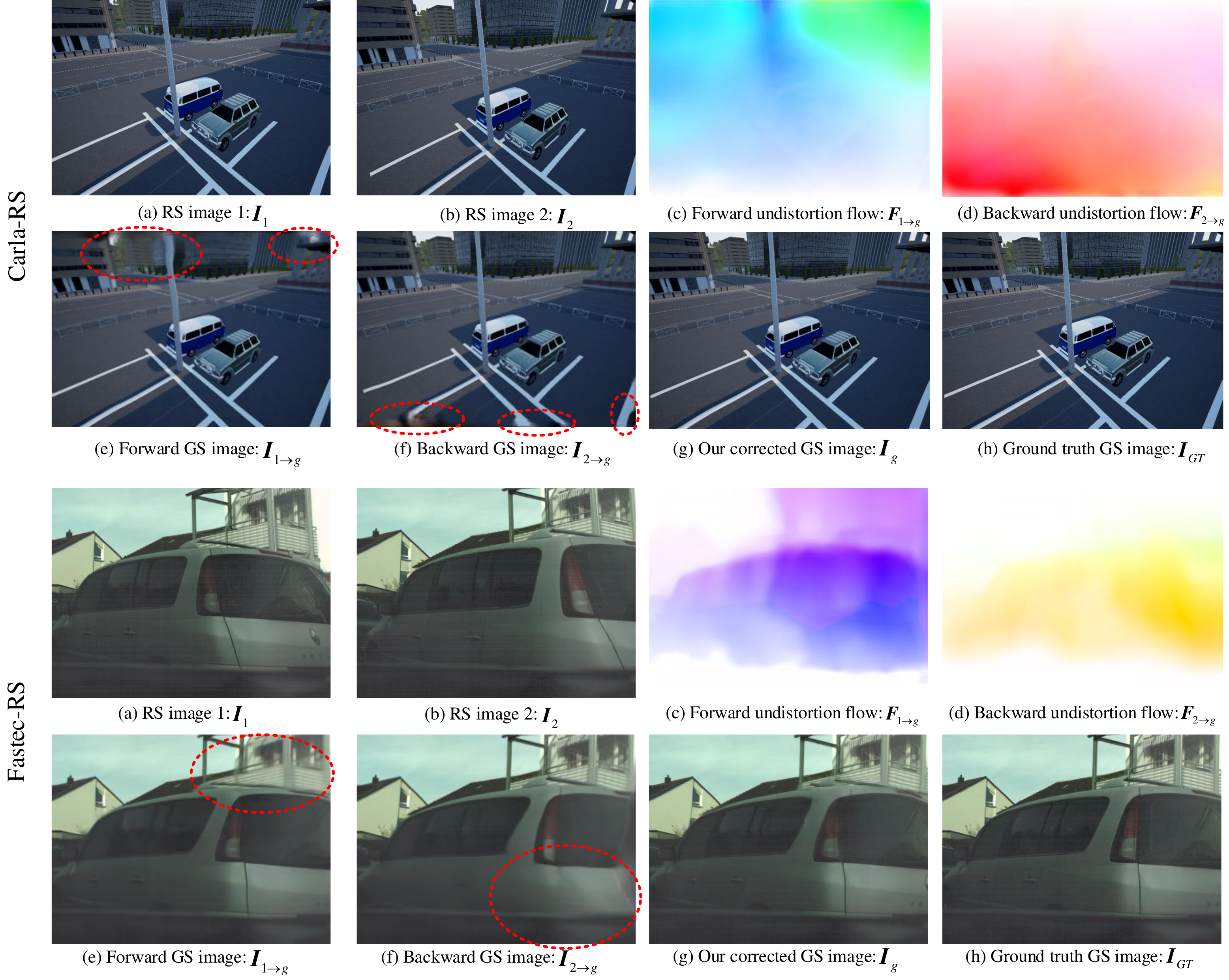}
	\caption{Intermediate examples of our network, including forward/backward undistortion flows and GS images. The upper two rows and the lower two rows show examples of Carla-RS and Fastec-RS datasets, respectively. Inputting two consecutive RS images ${\bm I}_1$ and ${\bm I}_2$, our approach estimates the forward and backward undistortion flows ${\bm F}_{1\to g}$ and ${\bm F}_{2\to g}$ to predict the forward and backward GS images ${\bm I}_{1\to g}$ and ${\bm I}_{2\to g}$, which then are aggregated to produce a time-centered corrected GS image ${\bm I}_{g}$ as the ground truth GS image ${\bm I}_{GT}$. Note that the undistortion flow of a pixel closer to the intermediate time of two frames appears as a lighter color (i.e., smaller values), such as the last rows of (c) and the beginning rows of (d), which accounts for the basic scanline-dependent characteristics of the undistortion flows.\label{fig:intermediate_results}}
\end{figure*}

In Fig.~\ref{fig:overview}(e)\&(f), we have shown the forward and backward GS images ${\bm I}_{1 \to g}$ and ${\bm I}_{2 \to g}$ synchronously generated by our network.
To better prove that our pipeline can effectively aggregate the contextual cues of two consecutive RS images, we further report the intermediate results of our network in Fig.~\ref{fig:intermediate_results}, \emph{i.e.}, forward/backward undistortion flows and GS images.
The biggest motivation that utilizing contextual aggregation for RS correction is inspired by this obvious observation, \emph{i.e.}, the first RS image ${\bm I}_{1}$ and the second RS image ${\bm I}_{2}$ have different contributions to the upper and lower regions of the corresponding ground truth time-centered GS image ${\bm I}_{GT}$ respectively, which is exemplified in Fig.~\ref{fig:intermediate_results}(e)\&(f).
Through symmetric network design, our improved PWC-based architecture can obtain a plausible RS-aware undistortion flow (\emph{i.e.}, the closer the pixel to the intermediate time $\tau$, the smaller the value of the undistortion flow is generally.) for subsequent accurate RS correction.

\begin{figure*}[!t]
	\centering
	\includegraphics[width=0.99\textwidth]{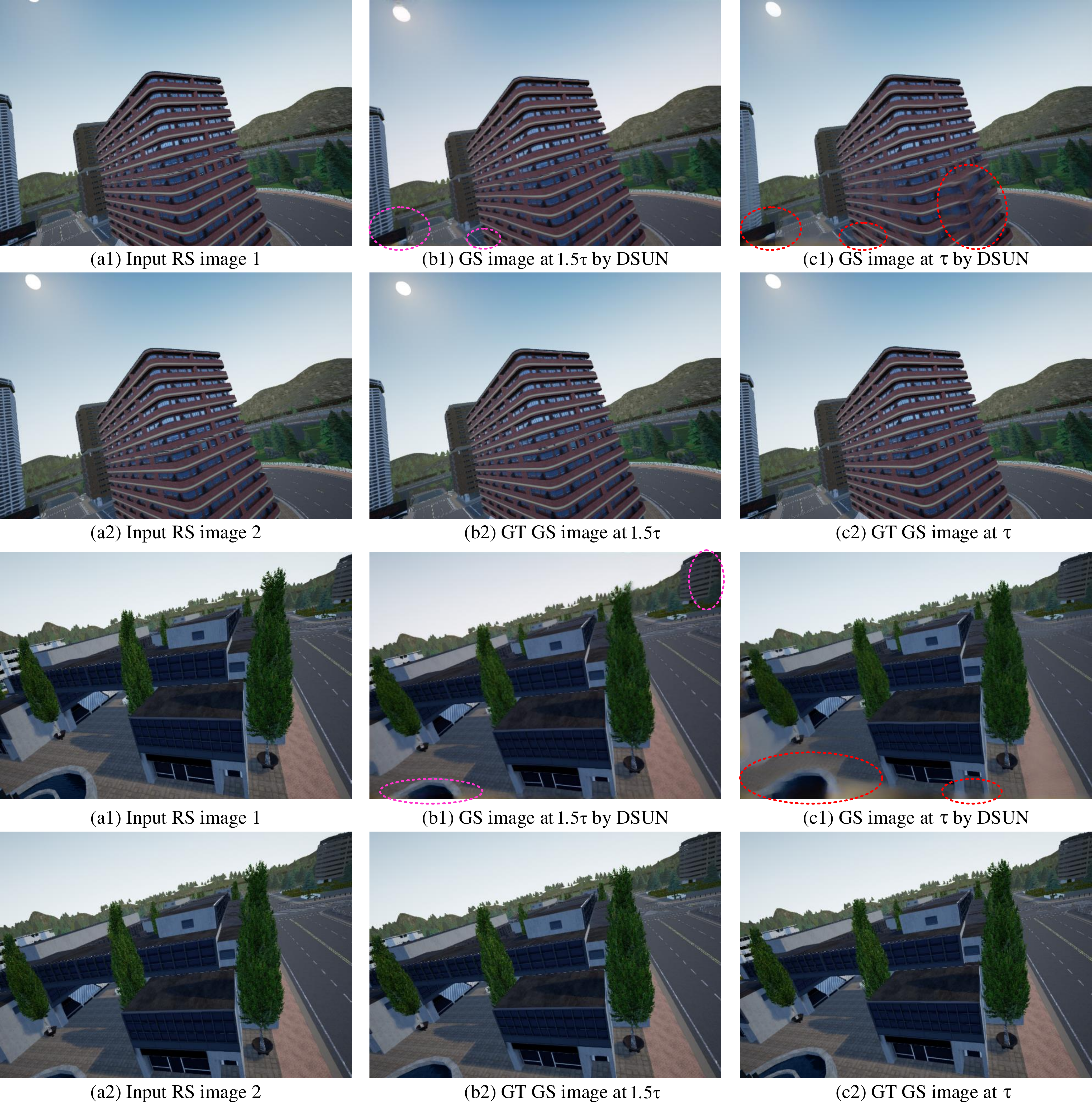}
	\caption{Two examples of vulnerability when generalizing DSUN \cite{liu2020deep} to estimate the time-centered GS image at time $\tau$. One can see that there are obvious differences between the latent GS images at $\frac{3\tau}{2}$ and $\tau$ in (b2) and (c2), and the GS image at $\frac{3\tau}{2}$ is relatively similar to the second RS image. DSUN cannot well recover the details of the GS image (c1) corresponding to time $\tau$ (see red circles), although it can obtain a plausible GS image (b1) corresponding to time $\frac{3\tau}{2}$ (In fact it also contains missing content that is not particularly striking, see pink circles). This is because the recovery of the GS image at time $\tau$ is more challenging than that at time $\frac{3\tau}{2}$. At this more challenging task, only the limited information of the second frame image is insufficient, and the context information must be fully utilized. Note that due to the effective context aggregation based on the imaging characteristics of consecutive RS images, our method can reconstruct the rich details of the latent GS image at time $\tau$, which is also the main contribution of our method.
		\label{fig:analysy_DSUN}}
\end{figure*}

As manifested in Fig.~\ref{fig:analysy_DSUN}, the limited information of the second RS image is insufficient to restore the rich texture details of the target GS image at intermediate time $\tau$. Intuitively, the rear of the car in the backward GS image marked by the red circle in Fig.~\ref{fig:intermediate_results}(f) is similar to the content restored by DSUN \cite{liu2020deep} in Fig.~\ref{fig:exp_Fastec_1}(d). This is because the extra information of the first frame is not used for detail extraction and fusion, which indicates the necessity of \textbf{contextual aggregation}.
Furthermore, our context-aware cost volume together with the symmetric consistency constraint is proven to be beneficial in effectively aggregating the contextual cues of two consecutive RS images, thereby resulting in high-quality time-centered GS images (at time $\tau$) with more complete visual content.

\section*{Appendix B: Network details}
\begin{figure*}[!t]
	\centering
	\begin{minipage}[t]{0.46\textwidth}
		\centering
		\includegraphics[width=4.08cm]{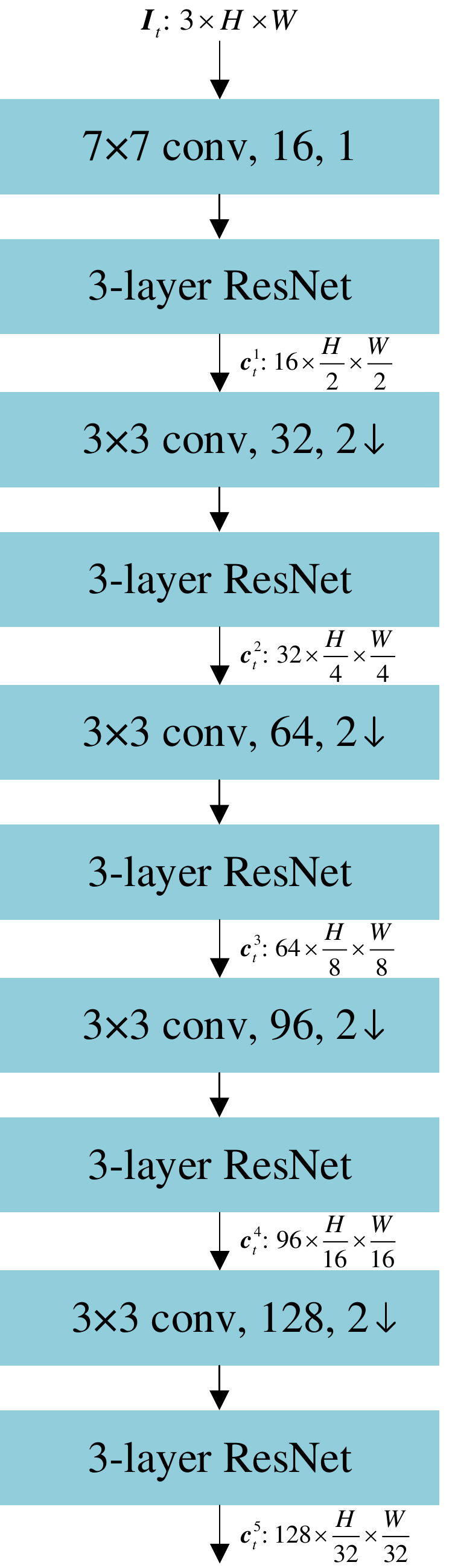}
		\caption{The feature pyramid extractor network. The first RS image $\bm{I}_1$ and the second RS image $\bm{I}_2$ are encoded using the same network. The convolutional layer and the $\times 2$ downsampling layer at each level is implemented using a single convolutional layer with a stride of 2, followed by a ReLU. Each ResNet layer contains two sequential blocks consisting of: a 2D convolution with a $3\times3$ kernel, a ReLU and a 2D convolution. $\bm{c}_t^l$ denotes extracted features of RS image $t$ at level $l$.\label{fig:feature_pyramid}}
	\end{minipage}
	\hspace{0.5cm}
	\begin{minipage}[t]{0.46\textwidth}
		\centering
		\includegraphics[width=6.38cm]{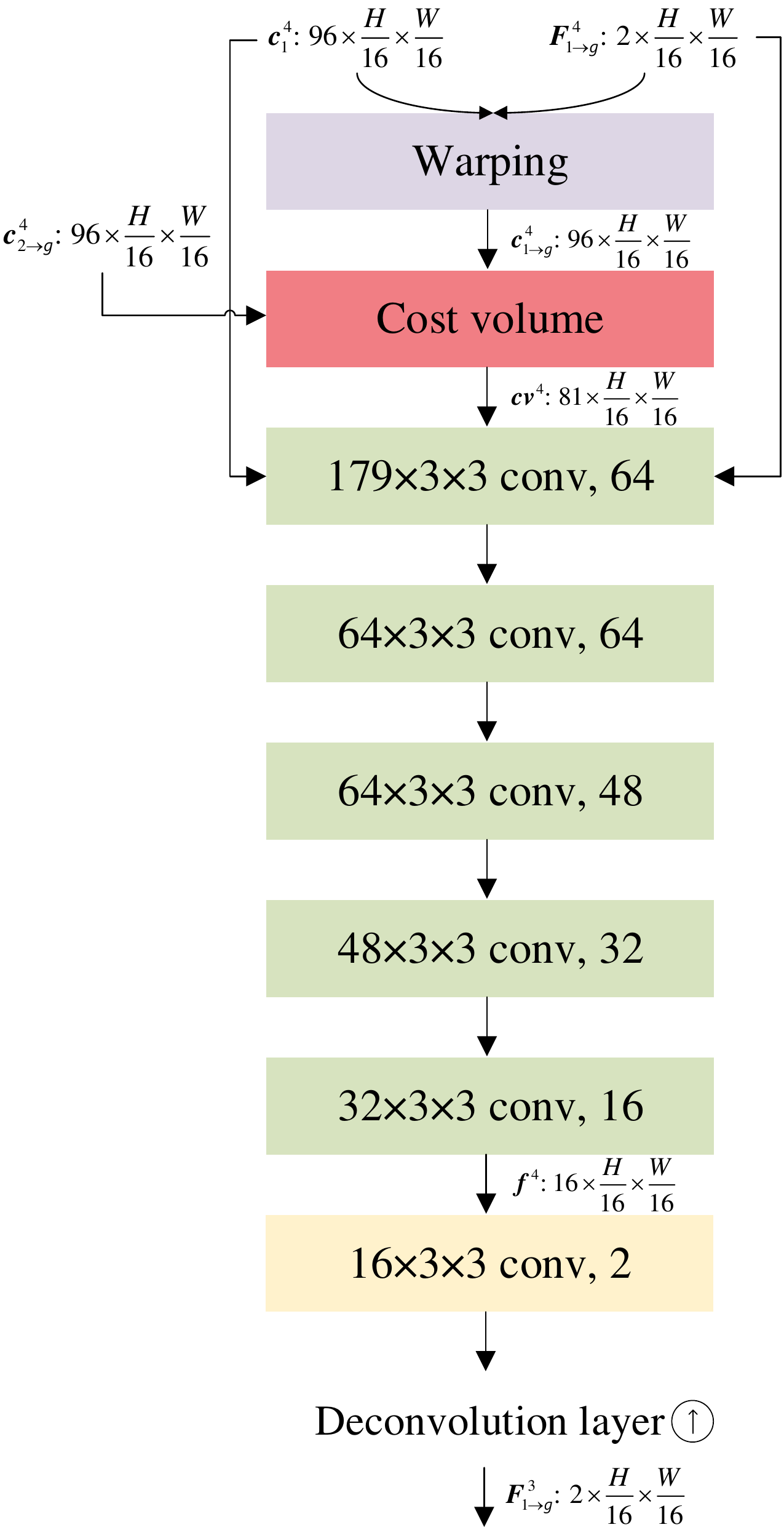}
		\caption{The undistortion flow estimator network of the first RS image $\bm{I}_1$ at pyramid level 4. Inspired by the DenseNet connections, each convolutional layer is followed by a ReLU except the last (yellow) one that outputs the undistortion flow. Deconvolution is then performed to return an upsampled undistortion flow $\bm{F}_{1 \to g}^3$ for subsequent processing.\label{fig:flow_estimator}}
	\end{minipage}
\end{figure*}

Fig.~\ref{fig:feature_pyramid} displays the architecture of the 6-level feature pyramid extractor network. Note that the bottom level indicates the original input RS images.
Fig.~\ref{fig:flow_estimator} illustrates the undistortion flow estimator network of ${\bm I}_1$ at the $4$-th pyramid level. The optical flow estimator networks at other levels have similar structures but different feature channels. 
Note also that the top level did not adopt the upsampled undistortion flows and calculated the cost volume using the pyramid features of the first and second RS images directly.


\section*{Acknowledgments}
This research was supported in part by National Key Research and Development Program of China (2018AAA0102803) and National Natural Science Foundation of China (61871325, 61901387).

{\small
	\bibliographystyle{ieee_fullname}
	\bibliography{SUNet_Arxiv}
}

\end{document}